\documentclass{ecai}
\usepackage[utf8]{inputenc}
\usepackage{amsthm}
\usepackage{amsmath}
\usepackage{mathtools}
\usepackage{tikz}
\usepackage{amsfonts}
\usetikzlibrary{shapes,arrows}
\usepackage[ruled,linesnumbered]{algorithm2e}
\usepackage{amsthm}
\usepackage{hyperref}
\usepackage{graphicx}
\usepackage{verbatim}
\newtheorem{remark}{Remark}
\begin{document}

\makeatletter
\def\blfootnote{\xdef\@thefnmark{*}\@footnotetext}
\makeatother

\title{A Normative approach to Attest Digital Discrimination}

\author{Natalia Criado, Xavier Ferrer, Jose M. Such \institute{Department of Informatics, King's College London,
UK, emails: \{natalia.criado,xavier.ferrer\_aran,jose.such\}@kcl.ac.uk} }

\maketitle
\bibliographystyle{ecai}

\begin{abstract}
Digital discrimination is a form of discrimination whereby users are automatically treated unfairly, unethically or just differently based on their personal data by a machine learning (ML) system. Examples of digital discrimination include low-income neighborhood’s targeted with high-interest loans or low credit scores, and women being undervalued by 21\% in online marketing. Recently, different techniques and tools have been proposed to detect biases that may lead to digital discrimination. These tools often require technical expertise to be executed and for their results to be interpreted.  To allow non-technical users to benefit from ML, simpler notions and concepts to represent and reason about digital discrimination are needed. In this paper, we use norms as an abstraction to represent different situations that may lead to digital discrimination. In particular, we formalise non-discrimination norms in the context of ML systems and propose an algorithm to check whether ML systems violate these norms.

\blfootnote{Author’s copy of the manuscript accepted in the Advancing Towards the SDGS Artificial Intelligence for a Fair, Just and Equitable World Workshop of the 24th European Conference on Artificial Intelligence (ECAI'20).}
\end{abstract}

\section{Introduction}
Digital discrimination is a form of discrimination in which automated decisions taken by algorithms, increasingly based on artificial intelligence techniques like machine learning, treat users unfairly, unethically, or just differently based on their personal data \cite{ours} such as income, education, gender, age, ethnicity, or religion. Digital discrimination is becoming a serious problem \cite{o2016weapons}, as more and more tasks are delegated to computers, mobile devices, and autonomous systems, for example some UK firms base their hiring decisions on automated algorithms.

Frequently the users of such machine learning (ML) systems are not technical experts and cannot assess by themselves if these algorithms are discriminatory. For example, many public organizations would like to reduce operational costs and delegate some decisions to algorithms, but at the same time need some guarantees about the ML systems not breaking anti-discrimination laws.  Our approach has been precisely designed to allow non-technical users to determine if ML systems are potentially discriminatory and to make explicit under which assumptions the systems are discrimination free. 

This paper is organised as follows: Section \ref{back} introduces background knowledge on discrimination legislation; Section \ref{norms} introduces our formalization of non-discrimination norms in the context of ML systems; Section \ref{alg} contains our attesting algorithm; Section \ref{case} illustrates the performance of our algorithm in two case studies; Section \ref{related} contains related work; and Section \ref{discussion}  contains a discussion of the paper contribution.

\section{Background\label{back}}
Many nation states and international organizations have enacted legislation prohibiting discrimination; e.g., the European Convention for the Protection of Human Rights.  
Most anti-discrimination legislation simply consists of a non-exhaustive list of criteria or protected attributes (e.g., race, gender, sexual orientation) on the basis of which discrimination is forbidden. Thus, discrimination are actions, procedures, etc., that disadvantage citizens based on their membership of particular protected groups defined by those attributes.

Legal systems traditionally distinguish between two main types of discrimination~\cite{altman2011discrimination}:
\begin{enumerate}
\item Direct discrimination (also known as disparate treatment) considers the situations in which an individual is treated differently because of their membership of a particular social group. This ultimately means that different social groups are being treated differently, with some of them effectively being disadvantaged by these differences in treatment. A clear example of direct discrimination would be a company having the policy of not hiring women with young children. 
Note, however, that direct discrimination does not necessarily involve that the discrimination process is explicit:
\begin{enumerate}
\item Direct discrimination can be explicit, exemplified in the previous case of a member of a particular social group (women with young children) explicitly disadvantaged by a decision process (hiring policy). 
\item Direct discrimination can be implicit, in situations in which the discriminated group is not explicitly mentioned. For example, the same company may replace the explicit hiring policy with a policy of not hiring candidates who have had a career break in recent years.
Although the policy does not explicitly refer to the relevant social group (women with children), it accomplishes the same task since woman with young children are statistically more likely to have had a recent career break.

\end{enumerate}
\item Indirect discrimination (also known as disparate impact) considers the situations in which an apparently neutral act has a disproportionately negative effect on the members of a particular social group. This is considered discrimination even if there is no intention to discriminate against that particular group or if there is not any unconscious prejudice motivating the discriminatory act. For example, a company having the policy to only consider customer satisfaction scores to award promotions may have a disproportionate impact on women, as there is empirical evidence suggesting that women are under evaluated when compared to their male counterparts with a similar objective performance.  In this case, the company may not have an intention to discriminate against female employees, but the promotion criteria set may effectively disadvantage them disproportionally.
\end{enumerate}
\section{Digital Discrimination Normative Model\label{norms}}

The term digital discrimination refers to those direct or indirect discriminatory acts that are based on the automatic decisions made by an ML system. In this section we formalise the notion of digital discrimination norms accounting for the different types of discrimination: explicit, implicit, and indirect. 

An ML system can be defined by a set of input features $\mathcal{I}=\{I_1,...I_m\}$, where each feature $I_i$ takes values from a discrete domain $D_{I_i}$; and an output feature $O$, which also takes values from a discrete domain $D_O$\footnote{For simplicity we assume domains are discrete, but any continuous domain can be discretized.}. Note in this paper we are interested in ML systems where the input contains personal information about individuals in order to attest discrimination. For this reason, the set of \emph{protected} features also needs to be defined; i.e., $\mathcal{P}=\{P_1,..,P_n\}$, where each protected feature $P_i\in\mathcal{P}$ takes values from a discrete domain $D_{P_i}$\footnote{Note that it is possible that protected features are part of the input used by a ML system, but is not necessary.}.

The decisions of an ML system can be represented as a dataset $DS$ formed by tuples $(p_1,...p_n,i_1,...i_m,o)$ representing a previous decision made by the ML system about a particular individual with protected attributes $p_1,...p_n$, input attributes $i_1,...i_m$, and algorithm outcome $o$\footnote{Note it is possible to consider the discrimination in an algorithm by considering also the ground-truth labels. See Appendix \ref{dalgoritm} for more details about this type of discrimination.}. In particular, each $p_i\in D_{P_i}$, $i_i\in D_{I_i}$ and $o\in D_{O}$. 

In the following, we provide a formalization of non-discrimination norms for ML systems and define how domain knowledge can be represented using norm exceptions. These normative notions are illustrated with an example.

\subsection{Digital Discrimination Norms}
As aforementioned, in the legislation we find the following types of discrimination: direct (also known as disparate treatment), which further classifies into explicit and implicit; and indirect (disparate impact) \cite{verma2018fairness}. In the following, we contextualise these notions in the context of digital discrimination and we formally represent them as computational norms using deontic logic\footnote{For simplicity, we don't consider compound discrimination. For a definition of compound discrimination norms see Appendix \ref{compound}}. 
\begin{itemize}
    \item Direct Discrimination: is the unequal behavior toward someone because of a protected characteristic. 
    We identify two types of direct discrimination:
    \begin{itemize}
        \item Explicit Discrimination. In terms of ML systems this equals to having some of the protected attributes considered in the algorithm input. Norms preventing explicit discrimination can be formalised as prohibitions to include protected attributes in the input of the system:
    \[\forall P_i\in \mathcal{P}: \mathbf{F}(P_i\in \mathcal{I})\]
    
    The set of all explicit discrimination norms is denoted by $N_E$ and has a size of $|\mathcal{P}|$.
        \item Implicit Discrimination can be formalised as a situation where the values of a set of input attributes correlate with the value of some protected attribute. 
        \[\forall P_i\in \mathcal{P}: \mathbf{F} (P_i \textrm{ is a function of }\mathcal{I})\]
        Note the fact that $P_i \textrm{ is a function of }\mathcal{I}$ needs to be defined in terms of a process to detect correlations or dependencies between attributes (Section \ref{related} provides more details about techniques that can be used for this).
        
        The set of all implicit discrimination norms is denoted by $N_I$ and has a size of $|\mathcal{P}|$ 
    \end{itemize}
    \item Indirect Discrimination (disparate impact) refers to decisions that adversely affect one group of people of a protected characteristic more than another. This equals to state that for a particular protected attribute value $p\in D_{P_i}$ the probability of a given outcome $o\in D_o$ is $x$ times lower than that of the values of the same protected attribute $p$ with the highest probability:
    \[
    \forall P_i\in \mathcal{P},  \forall p\in D_{P_i},  \forall o\in D_{O}: \mathbf{F}(P_i \downarrow_{o}^p)
    \]
    where $P_i \downarrow_{o}^p$ denotes:
    \[\mathit{Pr}(O=o|P_i=p)<x\times\max_{\forall p'\in D_{P_i}} \mathit{Pr}(O=o|P_i=p')\]
    $\mathit{Pr}(O=o|P_i=p)$ stands for probability that outcome $o$ is given to an individual with protected attribute $p$. Note that different methods can be used to estimate this probability, in Section \ref{related} we provide a review of  different techniques that can be used. The value $x\in[0,1]$ is a constant representing the disproportion allowed in a particular domain\footnote{For example the US \emph{fourth-fifth rule} from the Equal Employment Opportunity Commission (1978) states that a job selection rate for the protected group of less than 4/5 of the selection rate for the unprotected group \cite{Barocas2016}.}.

    The set of all disparate impact  norms is denoted by $N_D$ and has a size of $|\mathcal{P}|\times \overline{D_\mathcal{P}} \times |D_O|$, where $\overline{D_\mathcal{P}}$ denotes the average number of values belonging to the domain of protected attributes.
\end{itemize}
Discrimination norms are represented as a collection denoted by $N=(N_E,N_I,N_D)$.

\begin{remark}
If an explicit discrimination norm for a protected feature $P_i$ is violated, then the implicit discrimination norm for $P_i$ is also violated. The inverse inference is not true.
\label{implicationNorm}
\end{remark}

\begin{remark}
If an explicit discrimination norm for a protected feature $P_i$ is violated and no indirect discrimination norm for $P_i$ is violated, then the violation is inconsequential as the protected feature $P_i$ is not affecting significantly the decision-making process. If an implicit discrimination norm for protected feature $P_i$ is violated and no indirect discrimination norm for $P_i$ is violated, then the violation is inconsequential as the protected feature $P_i$ is not affecting significantly the decision-making process. \label{inconsequential}
\end{remark}
In this paper we will define inconsequential norm violations as those violations which can be considered trivial as they do have little effect on the decisions made by algorithms. Importantly, inconsequential violations are anyway worth considering as they may be an indicator of bad practice (e.g., considering disability status of students in university admissions may be immoral even if that information is not influencing much the decision).

\subsection{Norm Exceptions}
The previous section formalises the general definition of anti-discrimination norms. In general, when these norms are violated there is a potential case of digital discrimination. However, there are domains in which the violation of these norms is justifiable, and hence not result in discrimination. To allow for such type of domain knowledge to be explicitly represented and accounted for, we use the notion of domain permission norms, which define exceptions to the general anti-discrimination norms. 
\begin{itemize}
    \item Permission to use protected attributes in decision making. For example, legislation usually does not consider discriminatory to use religion as a criteria for hiring a religion teacher at a school. An explicit permission to use a protected attribute $P_i\in \mathcal{P}$ can be defined as follows:
    \[\mathcal{P}: \mathbf{P}(P_i\in I)\]
    The set of all exceptions to explicit discrimination norms is denoted by $E_E$.
    \item Permission to allow for correlations between a protected attribute and input attributes. For example, in some employments (e.g., firefighters) the employees should demonstrate physical strength, which is correlated with gender. In such cases, it is lawful to consider the results of fitness tests in hiring decisions.  This allowed correlation between a protected attribute $P_i\in\mathcal{P}$ and a subset of the input attributes $I\subset \mathcal{I}$ can be represented as a permission as follows:
        \[\mathbf{P}(P_i\text{ is a function of }I)\]
    The set of all exceptions to implicit discrimination norms is denoted by $E_I$.
    \item Permission to adversely affect one group. For example, on average women Uber drivers are paid less than men drivers \cite{cook2018gender}, but that is explained by factors such as driver experience, time and location of rides, etc. An exception to allow for a significant difference on an outcome $o\in D_o$ for a particular protected group $p \in D_{P_i}$ where $P_i\in \mathcal{P}$ can be formalised as follows: 
    \[
    \mathbf{P}(P_i\downarrow_{o}^{p})
    \]
    The set of all exceptions to indirect discrimination norms is denoted by $E_D$.
\end{itemize}
Domain exceptions to discrimination norms are represented as a collection denoted by $E=(E_E,E_I,E_D)$.

\begin{remark}
An exception to an explicit discrimination norm about protected attribute $P_i$ entails an exception for the implicit discrimination norm related to $P_i$ and all input attributes. The inverse relationship does not hold.
\label{implicationException}
\end{remark}

\begin{remark}
An exception to an explicit discrimination norm about protected attribute $P_i$ does not entail an exception to any indirect discrimination norms for $P_i$. 
An exception to an implicit discrimination norm about protected attribute $P_i$ does not entail an exception to any indirect discrimination norms for $P_i$. 
\end{remark}
There may be cases in which it is lawful to consider protected attributes in the decision-making process, either explicitly or implicitly, as long as that information is not used to disproportionately disadvantage the members of a certain group; e.g., positive discrimination practices may use gender information can be used to increase the number of employees from minority groups in a company or business, which are known to have been discriminated against in the past. In this case there is an exception to an explicit discrimination norm about gender, as long as that information is not used to adversely affect any group. 

%
%
\subsection{Example: Credit Risk Assessment}\label{sec:example}

To illustrate the different types of norms and exceptions let us consider an example of a decision making system that classifies people as high or low credit risks. 

In particular, the attributes used to describe people are:
$$\mathcal{I} = \{Age,  Job, Salary\}$$
\noindent  where and $Age \in \{[20,30],[30,40],...\}$, $Job\in \{Unemployed, \allowbreak Unskilled, ...\}$, and $Salary\in \{[0,20k],[20k,30k],...\}$. 
According to common discrimination law, protected attributes are defined as: $$\mathcal{P}=\{Gender, Age\}$$
where $Gender \in \{Male, \allowbreak Female\}$.
The output variable is:
$$O = Risk$$
where $Risk \in \{High, Low\}$.

In this example the following norms are generated considering protected attributes:
\begin{equation*}
\begin{split}
\mathbf{F}(Gender\in \mathcal{I}),& \mathbf{F}(Age\in \mathcal{I}),\\ \mathbf{F}(Gender\text{ is a function of } \mathcal{I}), &\mathbf{F}(Age\text{ is a function of } \mathcal{I}) \\ \mathbf{F}(Gender\downarrow_{High}^{Male}), &\mathbf{F}(Gender\downarrow_{Low}^{Male}),\\
\mathbf{F}(Gender\downarrow_{High}^{Female}), &\mathbf{F}(Gender\downarrow_{Low}^{Female}),\\
\mathbf{F}(Age\downarrow_{High}^{[20,30]}),&\mathbf{F}(Age\downarrow_{Low}^{[20,30]}),... \\
...,\mathbf{F}(Age\downarrow_{High}^{[70,80]}),&\mathbf{F}(Age\downarrow_{Low}^{[70,80]}),\\
\end{split}
\end{equation*}
However, in this example there are several exceptions to the norms:
\begin{equation*}
\begin{split}
\mathbf{P}(Age\in &\mathcal{I})\\ \mathbf{P}(Gender\text{ is a function of } &\{Salary\}),\\
\mathbf{P}(Age\downarrow_{High}^{[20,30]}),&\mathbf{P}(Age\downarrow_{Low}^{[20,30]}),...\\
...,\mathbf{P}(Age\downarrow_{High}^{[70,80]}),&\mathbf{P}(Age\downarrow_{Low}^{[70,80]}),\\ 
\end{split}
\end{equation*}
In particular, it is considered lawful to use age in credit risk assessment, as it is common practice to use age to estimate health, unemployment probabilities, etc. By Remark \ref{implicationException}, it is implicitly permitted that age is a function of input attributes. Moreover, it is considered permitted to allow age to have a significant impact on credit decisions. The pay gap phenomenon also explains a degree of correlation between salary and gender. In this case, however, the use of salary for credit risk assessment is lawful (i.e., salary has not been used as a way to discriminate women, but as a way to determine the capability of individuals to pay a credit back).

\section{Digital Discrimination Attesting Process\label{alg}}

\begin{figure}
\includegraphics[width=0.45\textwidth]{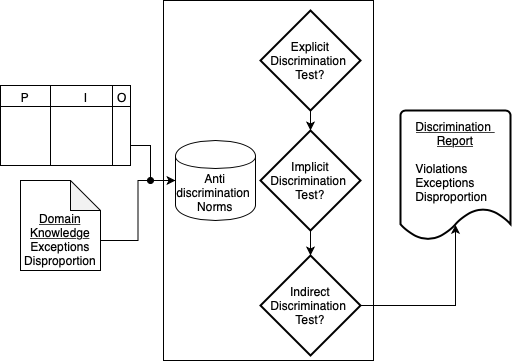}
\caption{Overview of the Attesting Process}
\label{Process}
\end{figure}

The digital discrimination attesting process (see Figure \ref{Process}) takes as input a dataset and the domain exceptions defined by the user, and it returns a discrimination report with information about any potential discrimination cases (i.e., a minimal list of norm violations) and about the assumptions made in the attesting process (i.e., the list exceptions provided by the user and the allowed disproportion ratio)\footnote{Note the purpose of our paper is to allow non-technical users to attest whether ML systems discriminate. For examples of research on mitigating discrimination see \cite{hajian2012methodology,calmon2017optimized,kamiran2012data,feldman2015certifying}.}.

The attesting algorithm (see Algorithm \ref{algorithm}) starts by generating the list of discrimination norms based on the input, protected and output attributes (line \ref{generation}) then it checks for:
\begin{itemize}
    \item Explicit direct discrimination (lines \ref{ES}-\ref{EE}). For each explicit norm that is violated a new inconsequential violation is added (later on the algorithm will confirm if this violation is inconsequential or not) and the implicit norm related to that protected attribute is removed. Note our goal is to produce the minimal set of violations and by Remark \ref{implicationNorm} the explicit norm is more general. 
    \item Implicit direct discrimination (lines \ref{IS}-\ref{IE}). For each implicit norm the algorithm checks if there is an exception to an explicit norm for the same protected attribute (as stated in Remark \ref{implicationException}). If not, the algorithm checks if the norm is violated using the dataset as a representative sample (line \ref{detectionIS})\footnote{Different statistical methods can be used to determine if there is a correlation between input attributes and protected attributes, see Section \ref{related} for more details.}. If the norm is violated, the algorithm checks for a permission allowing for that particular violation (lines \ref{ExS}-\ref{ExE}). An implicit norm is violated when there is a set of input attributes determining the value of a protected attribute.  In particular, the algorithm  checks if there is an exception for that set of input attributes, or a subset of it, determining the protect attribute. Again, if the norm is violated, a new inconsequential violation is created. 
    \item Indirect discrimination (lines \ref{DS}-\ref{DE}). For each indirect norm that is violated a new violation is created (line \ref{addV}). As stated in Remark \ref{inconsequential}, if there were some inconsequential violations related to that protected attribute, these are converted into consequential ones (lines \ref{rms}-\ref{rme}). The violation of the indirect norm associated to a protected attribute demonstrates that that decisions are having a disproportionate impact based on that protected attribute. 
\end{itemize}
The algorithm outputs the list of inconsequential and consequential violations. Note that the discrimination report will contain not only the information about norm violations (if any), but also the information about the exceptions considered in this analysis and the level of allowed disproportion specified by the user. 
 
\begin{algorithm}
\SetAlgoNoLine
\DontPrintSemicolon
  \caption{Digital Discrimination Attesting}
  \label{algorithm}
  \SetKwInOut{Input}{inputs}
  \SetKwInOut{Output}{output}
  \SetKwProg{DiscrminationAttesting}{DiscriminationAttesting}{}{}

  \DiscrminationAttesting{$(\mathcal{P,I},O,DS,E,x)$}{
    \Input{A set of protect attributes $\mathcal{P}$\\ A set of input attributes $\mathcal{I}$\\An output attribute O, a dataset $DS$\\A collection of exceptions $(E_E,E_I,E_D)$\\
    $x\in[0,1]$ a constant representing the disproportion allowed }
    \Output{A collection of violated norms $(V_E,V_I,V_D)$\\
    A collection of norms that have been violated inconsequentially $(I_D,I_I)$}
    $V_E\gets \emptyset$\;
    $V_I\gets \emptyset$\;
    $V_D\gets \emptyset$\;
    $I_E\gets \emptyset$\;
    $I_I\gets \emptyset$\;
    $(N_E,N_I,N_D)\gets GenerateNorms(\mathcal{P},\mathcal{I},O)$\; \label{generation}
    \tcp{Attesting Explicit Discrimination}
     \ForEach{$\mathbf{F}(P_i\in \mathcal{I})\in N_E$}{\label{ES}
       \If{$\not\exists \mathbf{P}(P_i\in \mathcal{I})\in E_E$}{%
       \If{$P_i\in \mathcal{I}$}{%
         $I_E\gets I_E \cup \{\mathbf{F}(P_i\in \mathcal{I})\}$\;
         $N_I\gets N_I\setminus \{\mathbf{F}(P_i\textrm{ is a function of } \mathcal{I})\}$\;
       }
      }
      }\label{EE}
    \tcp{Attesting Implicit Discrimination}
    \ForEach{$\mathbf{F}(P_i\textrm{ is a function of } \mathcal{I})\in N_I$}{ \label{IS}\label{detectionIS}
      \If{$\not\exists \mathbf{P}(P_i\in \mathcal{I})\in E_E$}{%
      \ForEach{$I\subseteq \mathcal{I}: I \textrm{ is the minimal set }\allowbreak\textrm{such that } P_i\textrm{ is a function of } I$}{\label{ExS}
        \If{$\not\exists \mathbf{P}(P_i\textrm{ is a function of }I'): I\subseteq I'$}{
            $I_I\gets I_I \cup \{\mathbf{F}(P_i\textrm{ is a function of }\mathcal{I})\}$\;
          }\label{detectionIS} \label{ExE}
      }
      }
      
    }  \label{IE}
    \tcp{Attesting Indirect Discrimination}
        \ForEach{$\mathbf{F}(P_i\downarrow_o^p)\in N_D$}{\label{DS}
          \If{$\lnot\exists \mathbf{P}(P_i\downarrow_o^p)\in E_D$}{%
              \If{$\exists p'\in D_{P_i}:\frac{  \mathit{Pr}(O=o|P_i=p)}{\mathit{Pr}(O=o|P_i=p')}<x$}{%
                $V_D\gets V_D \cup \{\mathbf{F}(P_i\downarrow_o^p)\}$\;\label{addV}
                \If{$\mathbf{F}(P_i\in \mathcal{I})\in I_E$}{\label{rms}
                $I_E\gets I_E \setminus \{\mathbf{F}(P_i\in \mathcal{I})\}$\;
                $V_E\gets V_E \cup \{\mathbf{F}(P_i\in \mathcal{I})\}$\;}
                \If{$\mathbf{F}(P_i\textrm{ is a function of } \mathcal{I})\in I_I$}{
                $I_I\gets I_I \setminus \{\mathbf{F}(P_i\textrm{ is a function of } \mathcal{I})\}$\;
                $V_I\gets V_I \cup \{\mathbf{F}(P_i\textrm{ is a function of } \mathcal{I})\}$\;}\label{rme}
          }
        }
        
    }\label{DE}
    $V\gets(V_E,V_I,V_D)$\;
    $I\gets(I_E,I_I)$\;
    \KwRet{V,I}\;
  
}

\end{algorithm}

The attesting algorithm complexity is determined by the size of the biggest norm set (or exception set). In this case, the complexity is given by  $\mathbf{O}(|\mathcal{P}|\times \overline{D_\mathcal{P}}\times |D_O|)$. This assumes that the norm violation checks are performed offline and can be retrieved in constant time. Section \ref{related} discusses different methods to check  compliance of implicit and indirect norms (note checking compliance of explicit norms equates to checking set membership).

\section{Case studies}\label{case}
In this section, we illustrate the performance of our digital discrimination attesting algorithm using two well-known datasets: the German dataset\footnote{\url{https://archive.ics.uci.edu/ml/datasets/statlog+(german+credit+data)}} and the Adult dataset\footnote{\url{https://archive.ics.uci.edu/ml/datasets/adult}}.

In our implementation\footnote{Available on Github at \url{https://github.com/xfold/NormativeApproachToDiscrimination}}, we have used the \emph{sklearn} library for normalised mutual information \cite{cover2012elements} to detect violations of implicit discrimination norms.
The normalised mutual information (NMI) is a measure of the mutual dependence between the two variables that quantifies the "amount of information" obtained about one random variable through observing the other random variable. The NMI returns 0 when there is no mutual information between the variables tested, and 1 when there exist a perfect correlation. In the implementation, the minimum coefficient for mutual information can be configured; we used a minimum threshold of 0.6 in the experiments below as indicative of a strong correlation between input and protected attributes. To detect indirect discrimination we have set to 0.8 the allowed disproportion ratio, inspired by the US \emph{fourth-fifth} rule from the Equal Employment Opportunity Commission (1978), a threshold commonly used to detect disparate impact in domains like employee selection procedures\footnote{\url{http://www.uniformguidelines.com}}. Also, due to the small size of the datasets used in the case studies, we have used the Chi-Squared Test \cite{cochran1952chi2} to determine those violations of indirect discrimination norms that are statistically significant (p-value $< 0.05$). To discretise numeric values, we have used quantile discretisation, which is a well-known method for discretising continuous variables in ML \cite{freese1967elementary}.

\subsection{Adult Dataset}
The Adult dataset uses 14 attributes to determine if a given person makes over 50K a year. The attributes include education, work class, age, sex, race, and occupation, among others.
The dataset contains 48842 instances.

Let us assume that the gender, age, native country and race are protected and that the other attributes are the inputs of a ML system. 
\begin{multline*}
\mathcal{I}=\{workclass, education, education\_num, occupation, \\
capital\_gain, capital\_loss, hours\_per\_week, fnlwgt\}
\end{multline*}
\begin{multline*}
\mathcal{P}=\{age, gender,native\_country,relationship,\\
marital\_status, race\}
\end{multline*}
\[O=income\]
where $income=\{<=50k,>50k\}$.
In this case age is related to experience and seniority so it is considered lawful to use age to discriminate:
\[\mathbf{P}(age\downarrow^{[0,16)}_{<=50k}), \mathbf{P}(age\downarrow^{[0,16)}_{>50k}), \] 
\[...\] 
\[\mathbf{P}(age\downarrow^{[75,99)}_{<=50k}),\mathbf{P}(age\downarrow^{[75,99)}_{>50k})\] 

After executing our algorithm several violations of indirect discrimination norms are detected. For example:
\[\mathbf{F}\ gender\downarrow_{>50k}^{female}\]
\[\mathbf{F}\ race\downarrow_{>50k}^{black}\]
\[\mathbf{F}\ native\_country\downarrow_{>50k}^{Nicaragua}\]
\[\mathbf{F}\ marital\_status\downarrow_{<=50k}^{Married-civ-spouse}\]
The violations above indicate that females, black people and nicaraguans have a disproportionate low probability of being classified as making more than 50k when compared with other groups, in accordance with previous reports of discrimination in the dataset \cite{bellamy2019ai}. On the contrary, married people are significantly less likely of being classified as making less than 50k. Found violations are associated with particular values of gender, native country, relationship and marital-status attributes. This indicates that the decision making process may have a disparate impact on people belonging to particular protected groups.

\subsection{German Credit Dataset}
The German dataset contains information about people who ask for a credit. Each person is classified as good or bad credit risks. This is the inspiration for the small example contained in section \ref{sec:example}. In particular, the full dataset uses 20 attributes to represent each person, which include information like 
age, employment status, gender and personal status of the applicant; and the duration, amount and purpose of the credit. 
The dataset contains 1000 instances.

Let's us assume an ML system where age, personal status and sex, and being a foreign worker are considered protected attribues, and the rest of the features in the German dataset are considered inputs: 
\[\mathcal{I}=\{job,housing,savings,..,amount,duration,purpose\}\]
\[\mathcal{P}=\{age, personal\_status\_and\_sex, foreign\_worker\}\]
\[O=risk\]
where $risk=\{high,low\}$. In this case, it is considered lawful to use age to discriminate credit risks as people are less likely to repay credits as they become older, hence, we consider age as an exception:
\[\mathbf{P}(age\downarrow^{[0,16)}_{good}), \mathbf{P}(age\downarrow^{[0,16)}_{bad}), \] 
\[...\] 
\[\mathbf{P}(age\downarrow^{[75,99)}_{good}),\mathbf{P}(age\downarrow^{[75,99)}_{bad})\] 
After executing our algorithm,  the following violation is detected: 
\[\mathbf{F}(foreign\_worker\downarrow_{good}^{yes})\]
The violation means that foreign workers have a disproportionate low probability of being considered a good credit risk.

\section{Related work}\label{related}
Recent research has addressed the problem of discrimination and bias in machine learning. These novel tools are most of the time aimed at technical users capable of interpreting different statistical results, programming, etc. Our algorithm is, on the contrary, aimed at non-technical users (albeit they may be domain experts). The notion of norm and exception is a suitable abstraction to represent the results these statistical analysis to non-technical users. 
For example, IBM's AI Fairness 360 Open Source Toolkit\footnote{\url{https://aif360.mybluemix.net}} 
and Google's What-if-tool\footnote{\url{https://pair-code.github.io/what-if-tool/}}, are probably two of the most comprehensive toolkits offering a great choice of bias metrics. However, its intended audience are technical users with previous knowledge of machine learning and statistics. Indeed, there are a large number of fairness metrics that may be appropriate for a given application \cite{verma2018fairness}. Also it is difficult for non-technical users to represent domain knowledge in a way that it can be taken into account by the metrics.  

Closely related to our work is \cite{10.1145/1568234.1568252}, where the authors proposed to infer classification rules from a given dataset and to detect those classification rules that can cause direct and indirect discrimination. They also allow for domain knowledge, expressed as rules, to be taken into account. Despite the similarities with this work, our proposal has 2 potential benefits: it doesn't assume that meaningful rules can be inferred, note that it may be impossible to infer rules from complex decision-making algorithms; and it hides to the user the complexities of the analysis process using the notion of norm and exception.

\paragraph{Implicit Discrimination. }
Tramèr et al. \cite{tramer2017fairtest} developed a methodology and toolkit combining different metrics for discovering associations, or proxies in observational data. 
In particular, they studied different metrics that can be used to analyse the relationship between protected attributes and input attributes such as the Pearson correlation, which only works for scalar attributes linearly related; and Mutual Information, which can be applied to categorical attributes. 
\paragraph{Indirect Discrimination. }
Within this line of research, \cite{zliobaite2015survey} surveys different metrics that have been proposed to measure indirect discrimination in data and the decisions made by algorithms. The study also discusses other traditional statistical measures that could be applied to measure discrimination. In particular, discrimination measures are classified by the authors into: statistical tests, which indicate the presence of discrimination at dataset level; absolute measures, which measure the magnitude of the discrimination present in a dataset; conditional measures, which capture the extent to which the differences between groups are due to protected attributes or other characteristics of individuals; and structural measures, which identify for each individual in the dataset if they are discriminated. In \cite{datta2016algorithmic}, the authors proposed quantitative metrics to determine the degree of influence of inputs on outputs of decision making systems. Their paper is not primarily intended to detect indirect discrimination, but the measures they propose have the potential to increase transparency of decisions made by opaque machine learning algorithms, which, in turn, may provide useful information for the detection of discrimination.

In addition to this work, there is  work also focusing on ML fairness. For instance, in \cite{dwork2012fairness}, they test for fairness based on a similarity measure between individuals. For fairness to hold, the distance between the distributions of outputs for individuals should at most be the distance between the two individuals as estimated by means of the similarity metric. 
In \cite{Grgic-Hlaca2018}, the authors first gather human judgments about the different protected features in the context of two real-world scenarios using Amazon Mechanical Turk. Using the set of \emph{human-assessed} protected features, they compare the accuracy of different classifiers to test the trade-off between process fairness and output accuracy. In \cite{kilbertus2017avoiding}, they assume fairness can be attested by means of a directed causal graph, in which attributes are presented as nodes joined by edges which, by means of equations, represent the relations between attributes.
Finally, the set of violations presented in our approach could also be extended with recent advances in explainable AI. One example is the post-hoc approach of Local Interpretable Model-Agnostic Explanations (LIME), which makes use of adversarial learning to generate counterfactual explanations \cite{Mueller2019}.

\section{Conclusion\label{discussion}}
Digital discrimination is becoming a significant problem as more decisions are delegated to ML systems. Indeed, recent legislation and citizen initiatives are demanding more transparency about the way in which decisions are made using their data. In response to that, several metrics and tools have been proposed to analyse biases in ML systems. However, these tools often require expert ML or statistical knowledge that many users of ML systems do not necessarily possess. 

In this paper, we proposed to use normative notions as an abstraction that may be more easily understood by non-technical users; simplifying the representation of the potential discrimination risks and the input of domain knowledge. Our digital discrimination attesting algorithm not only checks if ML systems are potentially discriminatory but also makes explicit under which assumptions these systems are discrimination free. 

As future work, we plan to: i) investigate different metrics to be used in the attesting algorithm and to identify the most usable ones; ii) conduct user studies to further refine the way in which norms could be accessed and influenced by non-technical users to help them understand discrimination risks.

\section*{Acknowledgements}
This  work  was  supported  by  EPSRC  under grant  EP/R033188/1.  It  is  part  of  the  Discovering and Attesting Digital Discrimination (DADD) project – see \url{https://dadd-project.org}.

\bibliography{dadd}

\appendix
\section{Compound Discrimination}\label{compound}
Compound discrimination is discrimination based on a combination of protected attributes. In that case of compound discrimination the previous discrimination norms are rewritten as follows:
\begin{itemize}
    \item Direct. 
    \begin{itemize}
        \item Explicit. There is no need to change the definition of explicit discrimination norms to account for compound discrimination, since the prohibition to include a set of protected attributes in the input can be represented by a set of explicit norms referring to each individual protected attribute. 
        \item Implicit. There is no need to change the definition of implicit discrimination norms to account for compound discrimination, since the prohibition to have a set of protected attributes as a function of input attributes can be represented by a set of implicit norms referring to each individual protected attribute. 
    \end{itemize}
    \item Indirect (disparate impact). In this case the norms need to represent that for a particular combination of protected attribute values ${p_1,...,p_k}$, where each $p_i\in P_i$; the probability of a given outcome $o\in D_o$ is $x$ times lower than for values of the same protected attributes with the highest probability:
\begin{equation*}
\begin{split}
    \forall \{P_1,...,P_k\}\subseteq \mathcal{P}, (p_1,...,p_k)\in& D_{P_1}\times...\times D_{P_k} , o\in D_{O}: \\\mathbf{F}(\{P_1,...,P_k\}& \downarrow_{o}^{(p_1,...,p_k)})
\end{split}
\end{equation*}
    where $\{P_1,...,P_k\}\downarrow_{o}^{(p_1,...,p_k)}$ denotes:
    
\scalebox{0.8}{
$\mathit{Pr}(O=o|P_1=p_1,...,P_k=p_k)<x\times\smashoperator{\max_{\forall \{(p'_1,...,p'_k)\}\in D_{P_1}\times...\times D_{P_k}}} \mathit{Pr}(O=o|P_1=p'_1,...,P_k=p'_k)
$}
 \end{itemize}
\section{Discrimination in Classification Process}\label{dalgoritm}
In this paper we have focused on digital discrimination; i.e., discriminatory acts facilitated by the automatic decisions made by a ML system. However, it is possible to consider the discrimination in the algorithm itself. In those cases it is necessary consider not only the outcome of the algorithm but also the ground-truth labels for the individuals, denoted by $G$. In those cases, it could be possible to formalise that for no particular value of a protected attribute the ML system can perform significantly worse than for the others groups
\[\forall P_i\in \mathcal{P}, p\in D_{P_i}, g\in D_{G}: \mathbf{F}(P_i\uparrow^p_g)\]
where $P_i\uparrow^p_g$ represents:
    \[\mathit{Pr}(O=g|P_i=p,G=g)<x\times\max_{\forall p'\in D_{P_i}} \mathit{Pr}(O=g|P_i=p',G=g)\]$\mathit{Pr}(O=g|P_i=p, G=g)$ stands for probability that the algorithm outcome $O$ is equal to the ground-truth label $g$ for an individual with protected attribute $P_i=p$.

\end{document}